# Inference for Multiplicative Models

Ydo Wexler          Christopher Meek

Microsoft Research, Microsoft Corporation
Redmond, WA 98052, USA

## Abstract

The paper introduces a generalization for known probabilistic models such as log-linear and graphical models, called here multiplicative models. These models, that express probabilities via product of parameters are shown to capture multiple forms of contextual independence between variables, including decision graphs and noisy-OR functions. An inference algorithm for multiplicative models is provided and its correctness is proved. The complexity analysis of the inference algorithm uses a more refined parameter than the tree-width of the underlying graph, and shows the computational cost does not exceed that of the variable elimination algorithm in graphical models. The paper ends with examples where using the new models and algorithm is computationally beneficial.

## 1 Introduction

Probabilistic models that represent associations and/or interactions among random variables have been heavily applied in the past century in various fields of science and engineering. The statistical methods originating with the work of Fisher (1925, 1956) [6, 7] culminated in the log-linear models which describe the association patterns among a set of categorical variables without specifying any variable as a response (dependent) variable [1].

A specific type of probabilistic models, probabilistic graphical models, can be visually described as an interaction graph, and embody independence assumptions in the domain of interest [15]. Their main attraction is that the independences encoded in the structure of the model allow to indirectly specify the join distribution as a product of functions $\psi_i(D_i)$, each depends only on a limited set of variables $D_i$. Algorithms that compute the posterior distribution conditioned on evidence, called *inference* algorithms, exploit this structure, avoiding a direct computation of the join probabilities [5, 19]. The complexity of such algorithms depends on the topology of the model, and is exponential in the tree-width of the underlying graph.

The common distinction within graphical models is between undirected graphical models [15], a subset of log-linear models, where there are no restrictions on the functions $\psi$, and Bayesian networks (BNs) [19] in which every function is a conditional distribution $\psi_i(D_i) = P(X_i|\Pi_i)$ where $\Pi_i$ is the set of parent variables of $X_i$ in the model. Another type of probabilistic models that can be represented visually, called factor graphs, extends undirected graphical models and incorporates many of the desired properties of graphical modes [14].

Aside of the independences that are imposed by the model's structure, often there exist additional independences stemming from the specific values of the functions. These independences are not systematically exploited by the traditional inference algorithms, resulting in an unnecessary computational cost. For such non-structural independences we use the name context-specific independence (CSI), which was suggested in previous studies [2, 20]. We note that the term CSI takes here a more general meaning as it is not restricted to any specific type of independence.

Several studies have suggested changes in the traditional representation of graphical models in order to capture context-specific independences. These include similarity networks suggested by Heckerman (1991) [12], multinets (Geiger & Heckerman 1996) [9], asymmetric influence diagrams (Fung and Shachter 1990) [8], and structured representations of the functions $\psi$ based on decision trees (Boutilier et al. 1996 [2], Poole& Zhang 2003 [20]). Other studies resorted to revised representations for specific functions (e.g. Quickscore algorithm by Heckerman 1989 for noisy-OR functions [11]).

Although the new representations proved useful from an empirical view point, they lack the ability to encompass a wide variety of CSI. In addition, the theoretical complexity of inference using these representation remained a function only of the topology of the graph underlying the model.

In this paper we approach the problem of inference from a more general perspective. We introduce a set of models called multiplicative models in which the functions $\psi$ that account for the dependency of variables are in a multiplicative representation, where a value of an instance is a product over a set of parameters. We show that multiplicative models generalize over log-linear models, factor graphs, and graphical models. In addition, we show that multiplicative models can capture multiple forms of CSI, including CSIs captured via decision trees, decision graphs, and via noisy-OR functions. This leads to the question whether an inference algorithm that takes advantage of these independences can be constructed without additional cost. We provide such an algorithm, and show how different types of independences are utilized in this procedure to reduce the needed computations. The inference algorithm provided herein simplifies over the inference algorithm suggested by Poole & Zhang (2003) [20] when applied to Bayesian networks, by avoiding the use of tables and tables splitting operations. The more general nature of the algorithm also enables it to deal with different representations, and thus account for CSI that can not be represented by decision trees and decision graphs.

We prove the correctness of the inference procedure and give a new notion of complexity instead of the exponent of the tree-width which is commonly used to describe the complexity of inference in graphical models. The new time complexity is shown to be less than or equal to the standard complexity.

## 2 Multiplicative models

We propose a generalization of graphical models, factor graphs and log-linear models which represents the dependency of variables in the model via the notion of multiplicative models. In these models a value of an instance in the dependency function is a product over a specific set of parameters. The definition relies on the concept of a lattice. A lattice $(L, \trianglelefteq, \cap, \cup)$ is a partially ordered set (poset) with respect to some relation $\trianglelefteq$, in which for every two elements $l_1, l_2 \in L$ their least upper bound is denoted as $l_1 \cap l_2$ and their greatest lower bound is denoted as $l_1 \cup l_2$.

We usually use upper case letters to denote random variables and sets of random variables, and lower case letters to denote their values. For a variable $V$ we denote its domain, or the set of possible values it can get, by $dom(V)$. For a set of variables $D = \{V_i\}_{i=1}^n$, the notation $dom(D)$ corresponds to the cross product of the domains $dom(V_i)$, $i = 1, \ldots, n$.

Let $D = \{V_i\}_{i=1}^n$ be a set of $n$ multivalued variables, and let the function $\psi(D) : dom(D) \to \mathbb{R}$ specify the values in a full table for the set $D$, then the following is a definition for a mapping function of $D$.

**Definition 1 (Mapping function)** *A function $f$ is called a mapping function of $D$ with respect to the lattice $L$, if it is defined as $f : dom(Z) \to L$ for every $Z \subseteq D$, and maps partial instances $Z = z$ onto $L$.*

We use this definition to define a lattice multiplicative model of $\psi(D)$.

**Definition 2 (Lattice multiplicative model)** *A model $\rho = \{S_\rho, \Gamma_\rho\}$ of a function $\psi(D)$ is called a lattice multiplicative model with respect to a lattice $(L, \trianglelefteq, \cap, \cup)$ and a mapping function $f$, if $S_\rho \subseteq L$, $\Gamma_\rho = \{\gamma_s \in \mathbb{R} : s \in S_\rho\}$ and $\psi(D = d) = \prod_{s \trianglelefteq f(d), s \in S_\rho} \gamma_s$.*

The set $S_\rho$ is called the structure of the model, and the set $\Gamma_\rho$ is called the parameters of the model. In multiplicative models elements $s \in S$ for which $\gamma_s = 1$ can be removed from $S$.

Here we focus on a lattice $L$ which is a set of propositional clauses over the variables and their values, and call this model a propositional multiplicative model, or simply a multiplicative model. In this model, the operators on the lattice are $\wedge$ and $\vee$. The mapping function used for this model is called the propositional mapping function, and is defined as follows.

**Definition 3 (Propositional mapping function)** *A mapping function $f$ is called a propositional mapping function of $D$ with respect to the lattice $L$, if for every set $Z \subseteq D$ the function maps every partial instance $Z = z$ into the conjunction $\bigwedge_{V_i \in Z} (V_i = v_i)$, where $v_i$ is the projection of $z$ onto the variable $V_i$.*

**Definition 4 (Propositional multiplicative model)** *A lattice multiplicative model $\rho = \{S_\rho, \Gamma_\rho\}$ of a function $\psi(D)$ is called a propositional multiplicative model with respect to a lattice $(L, \preceq, \wedge, \vee)$ and a propositional mapping function $f$, if the elements of $L$ are propositional clauses over the variables in $D$ and for two clauses $c$ and $c'$ we denote $c \preceq c'$ if $c$ is implied by $c'$.*

**Example 1** *Consider a set $D$ which contains two ternary variables $A$ and $B$. The corresponding lattice*

*contains propositional clauses over A and B, and for the two clauses $c = (A = 0)$ and $c' = (A = 0) \land (B = 2)$ we denote $c \preceq c'$. The corresponding mapping function maps the instance $A = 0, B = 2$ into the propositional clause $(A = 0) \land (B = 2)$, and the partial instance $A = 0$ into the clause $(A = 0)$.*

In this definition, the standard model which uses full-table representations of the functions $\psi(D)$, such as graphical models, and handles each instance separately, is also a multiplicative model with the set $S$ containing all mapping $f(d)$ of instances $D = d$, and with values $\gamma_d = \psi(d)$.

Another well-known model that falls into Definition 2 is the log-linear model.

### 2.1 Log-linear models

Log-linear models are usually used to analyze categorical data, and are a direct generalization of undirected graphical models. These models that have been heavily used for statistical analysis for the past four decades describe the association patterns among a set of categorical variables without specifying any variable as a response (dependent) variable, treating all variables symmetrically [1].

Formally, a log-linear model specifies the natural log of the expected frequency of values $d$ for a set of variables $D$ as a linear combination of the main effect $\lambda_{v_i}^{V_i}$ of every variable $V_i \in D$, and if $|D| > 1$ interaction effects $\lambda_s^S$ of every subset of variables $S \subseteq D$, where the instances $s$ are consistent with $d$. For example, suppose that we want to investigate relationships between three categorical variables, $A$, $B$ and $C$, then the full log-linear model is

$$\ln(F_{a,b,c}) = \mu + \lambda_a^A + \lambda_b^B + \lambda_c^C + \lambda_{ab}^{AB} + \lambda_{ac}^{AC} + \lambda_{bc}^{BC} + \lambda_{abc}^{ABC}$$

where $\mu$ is the overall mean of the natural log of the expected frequencies.

Clearly in the log-linear models instances are partially ordered by inclusion of their sets and by consistency of instantiations. To formalize log-linear models as a multiplicative models, for every subset $Z \subseteq D$ and for every instantiation $Z = z$ such that $\lambda_z^Z \neq 0$, the set $S$ contains all clauses of the form $\bigwedge_{V \in Z}(V = v)$, where $v$ is the projection of $z$ onto the variable $V$. In addition, we set the parameters of the model to $\gamma_\top = e^\mu$ and $\gamma_{f(s)} = e^{\lambda_s^S}$.

### 2.2 Context-specific independence

With the introduction of graphical models and in particular Bayesian Networks (BNs), and the proof that inference in these models is NP-hard [4], several studies looked for further independences encoded in models that can potentially reduce the amount of work needed for inference [12, 9]. The notion of Context-Specific Independence (CSI) was then introduced by Smith et al. (1993) [23] and Boutilier et al. (1996) [2]. Context-specific independence corresponds to regularities within probabilistic models based on the values assigned in the model.

Formally, we say that the sets of variables $X$ and $Y$ are contextually independent in the context of $C = c$ given $Z$ if

$$P(X, Y | Z = z, C = c) = \qquad (1)$$
$$P(X | Z = z, C = c) \cdot P(Y | Z = z, C = c)$$

for every value $Z = z$. One aspect of this equation is that if $X$ and $Y$ are contextually independent given $Z$, then

$$P(X | Y = y_1, Z = z, C = c) = P(X | Y = y_2, Z = z, C = c) \qquad (2)$$

for any two values $y_1, y_2$ of $Y$, which appear as repetitive values in conditional probability tables, such as those used in BNs. These repetition which are the basis of compact representations like decision trees and graphs were exploited for inference in BN [2, 20].

Another kind of CSI which was exploited for enhanced inference in BNs is the independence in noisy-OR functions. A noisy-OR function is a conditional probability function of a binary effect variable $E$ given a set of $m$ binary cause variables $C = \{C_1, \ldots, C_m\}$. The conditional probabilities of the function are $P(E = 0 | C_1, \ldots, C_m) = c_0 \prod_{i:C_i=1} P(E = 0 | C_i)$, where $c_0$ is a constant, and the values $P(E = 0 | C_i)$ are some real numbers.

For any particular CSI of the sets of variables $X$ and $Y$ in the context $C = c$ given the set $Z$, as in Eq. 1, there exists a multiplicative model that captures this independence. Such a model is any multiplicative model where the structure does not contain elements $s$ that involve variables from $X$ and $Y$, such that there exists an instance $Z = z$ for which $s \land (Z = z) = \bot$ and $s \land (C = c) \neq \bot$.

We now define two types of multiplicative models that capture two different types of common CSIs.

#### 2.2.1 Positive models

Representing the dependency of variables using log-linear models has some desirable properties, such as being general while ensuring the existence of a maximum likelihood without enforcing dependencies to be strictly positive. However, in the representation discussed in Section 2.1 the log-linear models use more parameters than necessary [3, 13]. Take for example the log-linear model for two binary variables $A$ and $B$.

Assuming all possible effects exist, the corresponding log-linear model uses eight parameters rather than the four parameters in a standard representation as a full table: $\lambda_0^A, \lambda_1^A, \lambda_0^B, \lambda_1^B, \lambda_{00}^{AB}, \lambda_{01}^{AB}, \lambda_{10}^{AB}, \lambda_{11}^{AB}$.

Another representation of the log-linear models that accounts for these redundancies uses only parameters which involve non-zero instantiations of variables [10]. In the above example the only parameters used in this representation are: $\lambda_1^A, \lambda_1^B, \lambda_{11}^{AB}$. We describe this representation of log-linear models as a multiplicative model, which we call here the positive model.

**Definition 5 (Positive model)** *A positive model $\rho$ of a function $\psi(D)$ is a multiplicative model wrt to the lattice $(L, \preceq, \wedge, \vee)$ and a (propositional) mapping function $f$ in which $S_\rho$ contains only elements $s = f(z)$ where $Z \subseteq D$ and no variable in $Z = z$ is set to zero.*

Log-linear models, and thus positive models, are known to capture conditional and contextual independences [16].

**Example 2** *An example is a function $\psi$ over two binary variables $A$ and $B$ where $\psi(0,0) \cdot \psi(1,1) = \psi(0,1) \cdot \psi(1,0)$. This implies that $A$ is independent of $B$ and the function can be written as $\psi(A,B) = \psi(A) \cdot \psi(B)$. In the corresponding positive model the parameter $\gamma_{(A=1) \wedge (B=1)} = \frac{\psi(0,0) \cdot \psi(1,1)}{\psi(0,1) \cdot \psi(1,0)} = 1$. Thus, this independence is captured in the model.*

**Example 3** *In a more complex function with three binary variables $A, B$ and $C$, every pair of variables is independent whenever the third variable is set to zero. For this function the corresponding positive model assigns $\gamma_{(V=1) \wedge (U=1)} = 1$ for every pair of variables $V, U \in \{A, B, C\}$ and where $V \neq U$.*

#### 2.2.2 Decision trees and graphs as multiplicative models

Common structures for representing functions with contextual independence are decision trees (DTs) and decision graphs (DGs) [22, 18]. These structures capture contextual independences that are the result of repetitive values, as specified in Eq. 2. Several studies have used decision trees to enhance inference in graphical models [2, 20]. We show how DTs and DGs fall into the category of multiplicative models.

For a function $\psi(D)$ over a set of variables $D$, a decision tree $T$ that represents $\psi(D)$ is a tree with variables from $D$ at internal nodes and values from $\psi(D)$ at the leaves. Every edge from a variable $V$ to a child in $T$ corresponds to a different set of values $H \subseteq dom(V)$, and can be represented as a set of clauses $\bigvee_{v \in H}(V = v)$.
A value at the end of a path $p = v_1 \to v_2 \to \cdots \to v_m$, where $v_i$ is some value of $V_i$, equals to the value of $\psi(d = v_1 v_2 \cdots v_m v_{m+1} \cdots v_n)$, where $V_j = v_j$ for $m < j \leq n$ is any possible value of $V_j$. We note that in a decision tree every instance $D = d$ is mapped to a single path in the tree. An example of a decision tree that encodes a function over the variables $A, B, C, D$ is shown in Figure 1.

One can choose to use decision graphs [18] instead of decision trees. These are more compact structures that can encode for more distributions. For a function $\psi(D)$ over a set of variables $D$, a decision graph $G$ that represents $\psi(D)$ is a directed graph with sets of variables from $D$ at internal nodes and values from $\psi(D)$ at the leaves. Similar to decision trees, every edge from a set of variables $W$ to a child $Z$ corresponds to a different set of values $H \subseteq dom(W)$, and can be represented as a set of clauses $\bigvee_{w \in H}(W = w)$. A value at the end of a path $p$ equals to the value of $\psi(d)$, where $d$ is an instance of $D$ consistence with the sets of values encoded by $p$. Again, as in decision trees, we note that in a decision graph every instance $D = d$ is mapped to a single path in the graph.

**Definition 6 (Decision-graph model)** *A decision graph model $\rho$ of a function $\psi(D)$ is a multiplicative model wrt to the lattice $(L, \preceq, \wedge, \vee)$ and a mapping function $f$ where every two elements $s_1, s_2 \in S_\rho$ satisfy $s_1 \wedge s_2 = \bot$, and $\bigvee_{s \in S} s = \top$, where $\bot = false$ and $\top = true$.*

For a specific decision graph $G$ that represents $\psi(D)$, the decision graph model of $G$ is $\rho(G)$ in which the structure contains one clause for every path from the root to a leaf in $G$, which is a conjunction of the clauses on the edges. For every such path $s$, we set $\gamma_s$ to the value at the end of the path. We note that in this model for every instance $D = d$ there is only one element $s \in S_\rho$ such that $s \preceq f(d)$.

## 3 Inference for multiplicative models

Consider a model that encodes for the probability distribution $P(x) = \prod_i \psi_i(d_i)$, with sets $D_i = \{X_{i_1}, \ldots X_{i_{m_i}}\}$, and multiplicative models $\rho_i = \{S_i, \Gamma_i\}$ over all the functions $\psi_i(D_i)$ wrt a lattice $(L, \preceq, \wedge, \vee)$. We first show how to perform inference, and compute a probability of a set of query variables $Q$ using a multiplicative model. In particular we perform inference for a multiplicative model via the variable elimination scheme (Zhang & Poole 1996 [24], Dechter 1999 [5]) which was originally suggested for inference in BNs. Then, we prove the correctness of the algorithm and analyze its time complexity.

We define an operation $M(V, \{\rho_i\})$, which given a variable $V \in X$ and a set of models $\{\rho_i\}, i = 1, \ldots, m$ over $X$ returns a model $\rho'$ over the variables $X \setminus V$. This operator is analogous to marginalization in standard inference algorithms. In addition, for a model $\rho$ we define a relevance indicator $I_s(V)$ for each element $s \in S_\rho$ and each variable $V$ in $D$, which is set to 1 if there exists a pair of instances $d_1, d_2$ of $D$ that differ only by the value of $V$ and for which $s \preceq f(d_1)$ but $s \not\preceq f(d_2)$. Otherwise, $I_s(V)$ is set to 0.

| A | B | C | D | $\psi(A,B,C,D)$ |
|---|---|---|---|---|
| 0 | 0 | 0 | 0 | 0.4 |
| 0 | 0 | 0 | 1 | 0.4 |
| 0 | 0 | 1 | 0 | 0.4 |
| 0 | 0 | 1 | 1 | 0.4 |
| 0 | 1 | 0 | 0 | 0.8 |
| 0 | 1 | 0 | 1 | 0.8 |
| 0 | 1 | 1 | 0 | 0.8 |
| 0 | 1 | 1 | 1 | 0.8 |
| 1 | 0 | 0 | 0 | 0.1 |
| 1 | 0 | 0 | 1 | 0.1 |
| 1 | 0 | 1 | 0 | 0.032 |
| 1 | 0 | 1 | 1 | 0.08 |
| 1 | 1 | 0 | 0 | 0.1 |
| 1 | 1 | 0 | 1 | 0.1 |
| 1 | 1 | 1 | 0 | 0.65 |
| 1 | 1 | 1 | 1 | 0.08 |

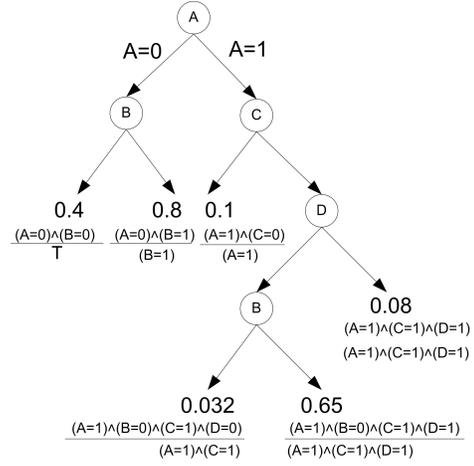

Figure 1: (left) A full-table over the binary variables $A, B, C, D$ that specifies the value of the function $\psi$ for each instance. (right) A decision tree corresponding to the function $\psi$ on left. Under every node in the tree appears the corresponding proposition in the decision-tree representation, and below the corresponding proposition in a positive representation of the propositions in the decision tree.

These operations allow us to write an inference procedure which computes the probability of a set of query variables in a multiplicative model as in Algorithm 1.

The algorithm operates like the bucket-elimination algorithm [5], where given an order on the variables we iterate over them (Line 2), and marginalize out one variable at a time (Line 9). Only elements that include terms that involve the current variable are considered in the marginalization.

Note that for graphical models, in which the elements of $S_i$ are a mapping of instances of the functions $D_i$, this algorithm is exactly the known variable elimination algorithm, in its implementation as bucket-elimination [5], where the sets $S_i[j]$ are the tables in the bucket of the variable $X_j$.

A general algorithm for computing $M(V, \{\rho_i\})$ is given as Algorithm 2. We use there the notation $s \lor V$ for an element $s \in S$ and a variable $V$ to denote $s \bigvee_{V=v} (V = v)$. This operation removes all terms that specify a value for $V$. For example, if $s = (V = 0) \land (U = 0)$ then $s \lor V = (U = 0)$.

The algorithm has two main parts: upto Line 5 the algorithm generates the set $R$ of possible new elements in the model. From Line 6 it computes the new parameters $\gamma_r$, where at each iteration a "minimal" element

---

**Algorithm 1**: VE for multiplicative models

**Input**: A model with $n$ variables $X_i$ ($i = 1, \ldots, n$) and $m$ functions $\psi_i(D_i \subseteq X)$, that encodes for the distribution $P(X)$. A set of multiplicative models $\rho_i = \{S_i, \Gamma_i\}$ wrt a mapping function $f$, where $\rho_i$ model $\psi_i(D_i)$, and a set of $k$ query variables $Q = \{X_i : i \leq k\}$

**Output**: The distribution $P(Q)$.

1   $t = m + 1$;
2   **for** $j = k+1$ *to* $n$ **do**
3     **for** $i = 1$ *to* $t-1$ **do**
4       $S_i[j] \leftarrow \{s : s \in S_i \,,\, I_s(X_j) = 1\}$;
5       $\Gamma_i[j] \leftarrow \{\gamma_s : s \in S_i[j], \gamma_s \in \Gamma_i\}$;
6       $S_i \leftarrow S_i \setminus S_i[j]$;
7       $\Gamma_i \leftarrow \Gamma_i \setminus \Gamma_i[j]$;
8     end for;
9     $\{S_t, \Gamma_t\} \leftarrow M(V, \{S_i[j], \Gamma_i[j]\})$;
10    $t = t + 1$;
11 end for;
12 $P(Q) \leftarrow \left\{ P(q) = \prod_{s_i \preceq f(q)} \gamma_{s_i} : Q = q, s_i \in S_i \right\}$;
13 return $P(Q)$;

---

Figure 2: Algorithm for variable elimination with a multiplicative model

of $R$ is chosen, and selects those elements $r$ with parameters $\gamma_r \neq 1$.

To compute the possible new elements, Lines 2 and 3 first create a closure under the operator $\land$ of each structure $S_i$. Then, in Line 5 all conjunctions of terms

**Algorithm 2**: $M(V, \{\rho_i\})$

**Input**: A variable $V$ and a set of representations $\rho_i = \{S_i, \Gamma_i\}$, $i = 1, \ldots, t$, wrt a lattice $(L, \preceq, \wedge, \vee)$, where $I_{s_i}(V) = 1$ for every $i$ and $s_i \in S_i$.
**Output**: A representation $\rho' = \{S', \Gamma'\}$.

1 $S' \leftarrow \emptyset$; $\Gamma' \leftarrow \emptyset$;
2 **for** $i = 1$ **to** $t$ **do**
3 $\quad R_i = \{\bigwedge_{s' \in S'_i} s' : S'_i \subseteq S_i\}$;
4 **end for**;
5 $R \leftarrow \{\bigwedge_{1 \leq i \leq t} r_i : r_i \in R_i\}$;
6 **while** $R \neq \emptyset$ **do**
7 $\quad$ * $r$ is a minimal element in $R$ *
8 $\quad r \in Min(R) = \{r' : r' \in R, \nexists r'' \in R \text{ s.t. } r'' \prec r'\}$;
9 $\quad \gamma_r = \dfrac{\sum_{V=v} \prod_i \prod_{s \preceq (r \wedge (V=v)), s \in S_i} \gamma_s}{\prod_{r' \in S', r' \preceq r} \gamma_{r'}}$;
10 $\quad R \leftarrow R \setminus \{r\}$;
11 $\quad$ **if** $\gamma_r \neq 1$ **then**
12 $\quad\quad S' \leftarrow S' \cup \{r\}$;
13 $\quad\quad \Gamma' \leftarrow \Gamma' \cup \{\gamma_r\}$;
14 **end while**;
15 **return** $\{S', \Gamma'\}$;

Figure 3: Algorithm for computing the operation $M(V, \{\rho_i\})$.

from the different closures consist of the set of possible new elements. In analogy to inference in graphical models, this operation is equivalent to the operation of tables' multiplication, often denoted as $\otimes$. In these models the set $R$ is the set of instances in the table after marginalization.

We note that for some models, like graphical models, lines 2-5 are trivial, and are executed implicitly, since the elements in $R$ are known to be all instances of a full-table over variables in $\bigcup S_i$.

### 3.1 Correctness of the inference procedure

We prove the correctness of Algorithm 1 by showing that the algorithm maintains the property that after iterating over the set of variables $U$, the models $\rho_i = \{S_i, \Gamma_i\}$ encode to the probability distribution $P(X \setminus U)$.

At the beginning of the algorithm every model $\rho_i$ represents the corresponding function $\psi_i(D_i)$. Thus,

$$P(X = x) = \prod_i \psi_i(D_i = d_i) = \prod_i \prod_{s \preceq f(d_i), s \in S_i} \gamma_{is}.$$

Assume that after removing the set of variable $U$ we are left with the set $X' = X \setminus U$, and now wish to eliminate a variable $V \in X'$. We write the probability of an instance $x'_v$ of $X' \setminus V$ which is the projection of an instance $X' = x'$ onto $X' \setminus V$ via the parameters $\gamma$:

$$P(x'_v) = \sum_{V=v} P(x') = \sum_{V=v} \prod_i \prod_{s \preceq f(x'), s \in S_i} \gamma_{is}$$

We can decompose the product into terms that involve the variable $V$ and those which do not. Denoting $\alpha(x'_v) = \prod_i \prod_{s \preceq f(x'), s \in S_i, I_s(V)=0} \gamma_{is}$, we get

$$P(x'_v) = \alpha(x'_v) \cdot \sum_{V=v} \prod_i \prod_{s \preceq f(x'), s \in S_i, I_s(V)=1} \gamma_{is}. \quad (3)$$

Now, lets examine what the algorithm encodes for after removing variable $V$, and show that it equals Eq. 3. While the elements that do not involve variable $V$ are not changed, the elements that do involve $V$ are removed and the elements $s_t \in S_t$ are added. Therefore, after applying Algorithm 2 for $V$ the remaining sets encode for $\hat{P}(x'_v) = \alpha(x'_v) \cdot \beta(x'_v)$ where $\beta(x'_v) = \prod_{s_t \preceq f(x'), s_t \in S_t} \gamma_{s_t}$. To express $\beta(x'_v)$ in the terms of Algorithm 2, recall that $S_t \subseteq R$ and if an element $s \in R$ and $s \notin S_t$ then $\gamma_s = 1$. Thus, we can rewrite $\beta(x'_v)$ using elements of $R$ as

$$\beta(x'_v) = \prod_{r \preceq f(x'), r \in R} \gamma_r.$$

From lines 2-5 in Algorithm 2, there is one element $r^* \preceq f(x')$ in $R$ for which $\forall r \in R$ such that $r \preceq f(x')$ also satisfies $r \preceq r^*$. First, to show there is such an element $r^*$ we recall from Line 5 that all elements in $R$ can be written as $r = \bigwedge_{1 \leq i \leq t} r_i$, where $r_i \in R_i$, and $R_i$ is the closure of $S_i$ under the operator $\wedge$. Consider the set of elements $r^*_i \preceq f(x')$, $i = 1, \ldots, t$, for which all other elements $r_i \in R_i$ such that $r_i \preceq f(x')$ satisfy $r_i \preceq r^*_i$. Then, every element $r = \bigwedge_{1 \leq i \leq t} r_i$ such that $r \preceq f(x')$ also satisfies $r \preceq r^*$.

Now, assume by contradiction that there were two such elements, $r^*_1, r^*_2 \in R$. Then from the definition of $r^*_1$ and $r^*_2$ we get $r^*_1 \preceq r^*_2$ and $r^*_2 \preceq r^*_1$, yielding $r^*_1 = r^*_2$. Thus, from line 9 in Algorithm 2

$$\beta(x'_v) = \gamma_{r^*} \cdot \prod_{r \preceq r^*, r \in R} \gamma_r = \sum_{V=v} \prod_i \prod_{s \preceq (r^* \wedge (V=v)), s \in S_i} \gamma_s$$

where the last equality is due to the fact that the denominator in the computation of $\gamma_{r^*}$ is $\prod_{r \preceq r^*, r \in R} \gamma_r$. In the terms of Algorithm 1 the set $\{s : s \preceq (r^* \wedge (V =$

$v))$, $s \in S_i\}$ can be rewritten as $\{s : s \preceq f(x'), s \in S_i, I_s(V) = 1\}$. Thus, we can write

$$\beta(x'_v) = \prod_i \prod_{s \preceq f(x'), s \in S_i, I_s(V)=1} \gamma_s$$

and from Eq. 3 we get $\hat{P}(x'_v) = P(x'_v)$. Namely, the new models encode for $P(X' \setminus V)$.

### 3.2 Incorporating evidence

In many practical scenarios we observe the value of some of the variables in the model, and wish to incorporate this evidence. The multiplicative models allow us to do so in a most natural way. Consider a set $E$ of evidence nodes for which we observed the values $E = e$, and a multiplicative model $\rho = \{S_\rho, \Gamma_\rho\}$. Then, in order to incorporate the evidence into $\rho$, we adjust the elements in $S_\rho$ by $s = s \bigwedge_{V \in E} (V = v)$, where $v$ is the projection of $e$ onto the variable $V \in E$. Then, we remove every element not consistent with the evidence, $s = \bot$.

### 3.3 Complexity of inference

It is well known that the complexity of inference in graphical models is NP-hard and its cost exponential in the tree-width of the underlying graph [4].

We analyze the time complexity of the inference procedure for multiplicative models given in Algorithm 1. As a by-product we refine the standard complexity and provide a new complexity bound which is based on the representation used. One can then say that the complexity of the problem is the minimum complexity among all possible representations.

#### 3.3.1 Diameter of multiplicative models

The structure of a multiplicative model determines the amount of computations needed to obtain the value $\psi(d)$ of a single instantiation of values to variables in a set $D$. Although at first glance it seems that for a model $\rho = \{S, \Gamma\}$ of a function $\psi(D)$ the number of operations needed to obtain values of all instances $D = d$ amounts to a total of $\sum_{D=d} |\{s : s \preceq d\}|$, the real number of operations can be dramatically lower and we denote it by $\delta(\rho)$. For hierarchical models, in which if an element $s$ is not in the structure of the model then all elements $s \preceq s'$ are also not in the model, Good (1963) provides a method that computes all such values in time $|S| \log |S|$ [10]. We denote the ratio between the number of computations and the number of elements in $S$, which is the size of the model, by $diam(\rho) = \frac{\delta(\rho)}{|S|}$ and name it the diameter of $\rho$.

From a computational perspective, it is clearly beneficial to use models with a small diameter, as this directly leads to fewer operations whenever we want to either obtain a value of $\psi$ or update the values $\gamma_s$. Examples of models with a diameter of 1 are graphical models and decision graph models, in which for every element $s \in S$, the only element $s'$ such that $s' \preceq s$, is $s$ itself. On the other hand, the diameter of a positive model can be as high as $\frac{\log |S|}{2}$. This maximum is achieved for a positive model of $m$ binary variables, when all $2^m$ parameters do not equal one, and hence all possible elements are in $S$. In this scenario the diameter is exactly $\frac{m}{2}$.

Although in the worst scenario the diameter of a positive model can be large, often this is not the case, and the diameter is typically bounded to be very small.

**Example 4** *Consider as an example the Potts model [21] in which a function $\psi(D)$ over a set $D = \{V_i\}_{i=1}^n$ decomposes according to $\psi(D = d) = c_0 \prod_{i,j} \psi(v_i, v_j)$, where $v_i$ and $v_j$ are projections of $d$ onto the variables $V_i$ and $V_j$ respectively, and $c_0$ is a constant. Although in general a positive model over $n$ binary variables has a diameter of $\frac{n}{2}$, in this example, the structure of the positive model includes only elements that involve at most two variables. Therefore, the diameter of the model is bounded by two.*

*Similarly, in a more complex scenario where the function $\psi$ decomposes to functions of $k$-tuples of variables, the diameter will be bounded by $k$.*

Consider a tree decomposition of the graph in which there is an edge between a pair of variables $V, U$ if there exists an element $s$ in one of the models for which $I_s(V) \cdot I_s(U) = 1$. We denote by $S(W) = \{s \vee (X \setminus Z) : s \in S_i\}$ the set of parts of elements in the models $\rho_i$ that involve variables from the set of graph vertices $Z$ which is mapped onto the tree node $W$. Further denoting as $S^-(W)$ the closure of $S(W)$ under the operator $\wedge$, we say that complexity of the algorithm for this tree decomposition is the maximum over the nodes $W$ in the tree of $|S^-(W)| \cdot diam(S^-(W))$, as described in Section 3.3.1. Then, the overall complexity of the algorithm is the complexity for the tree decomposition that yields the minimum for this term.

To see that this is indeed the time complexity of the algorithm, consider the elements in a set $R$ in Algorithm 2. The number of elements there does not exceed the number of elements in $S^-(W)$ for the corresponding tree decomposition and where $W$ maps onto the variables that appear in $R$. Most of the computation stems from computing the products in Line 9, and these can be done for the entire set of elements of $R$ in time proportional to $|R| \cdot diam(R)$. Therefore, having the ability to choose an elimination order, the complexity of the algorithm is $|S^-(W)| \cdot diam(S^-(W))$ maximized over all nodes $W$ in a tree decomposition and minimized over all possible such decompositions.

## 3.4 Benefits of inference for multiplicative models

Different multiplicative models capture different contextual independences, hence specifying different number of parameters. Take for example the function over four binary variables $A, B, C, D$ with values according to the table in Figure 1. The structure of the corresponding decision-tree model contains six elements while the structure of the corresponding positive model contains eight elements. In this latter model, the CSI captured in the decision tree, yielding the value of $\psi$ to be independent of $B$ given that $A, C$ and $D$ are set to one, does not have any effect. This variation and the structure of the model affect the run time of the inference algorithm.

An example where there are substantial computational savings when using the inference algorithm proposed can be found in a model such as the QMR-DT network [17], which is comprised of noisy-OR functions, mentioned in Section 2.2. The QMR-DT network is a two-level or bipartite BN where all variables are binary. The top level of the graph contains nodes for the diseases $C$, and the bottom level contains nodes for the findings $E$. The conditional probabilities in the network $P(E_i = e_i | \Pi_i)$, where $\Pi_i$ are the parents of finding $E_i$ in the network, are represented by noisy-OR functions.

Heckerman (1989) has developed an algorithm, called Quickscore, which takes advantage of the independence of the cause variables in the context of a negative finding $E_i = 0$ and uses it to speed up inference in the QMR-DT network [11].

For every noisy-OR function $P(E|C_1, \ldots, C_m)$ a structure of a multiplicative model that captures the independence does not contain elements $s$ such that $s \wedge (E = 0) \neq \bot$ for which $I_s(C_i) = 1$ and $I_s(C_j) = 1$, for all $i, j \leq m$.

In addition, running Algorithm 1 using multiplicative models with structures

$$S_i = \{(E_i = 1) \bigwedge_{C_i \in \Pi_i} (C_i = c_i) : \forall C_i = c_i\} \vee$$

$$\{(E_i = 0) \wedge (C_i = 1) : C_i \in \Pi_i\} \vee ((E_i = 0) \bigwedge_{C_i \in \Pi_i} (C_i = 0))$$

is identical to the Quickscore algorithm and gains the same savings automatically.